\documentclass[conference]{IEEEtran}
\IEEEoverridecommandlockouts

\usepackage{cite}
\usepackage{amsmath,amssymb,amsfonts}
\usepackage{graphicx}
\usepackage{textcomp}
\usepackage{xcolor}
\usepackage{hyperref}
\usepackage{listings}
\usepackage{url}
\usepackage{array}
\usepackage{booktabs}
\usepackage{float}
\usepackage{algorithm}
\usepackage{algorithmic}
\usepackage{multirow}

\begin{document}

\title{Autonomous Agricultural Tractor: Integrated Weed Detection and LiDAR Navigation for Precision Paddy Farming}

\author{
\IEEEauthorblockN{Benjamin Merryman-Smith, Tony Nguyen, Bilal Dogutas, Krish Shah, Anthony Raphael, Sudip Dhakal}
\IEEEauthorblockA{Department of Computing and Software Engineering\\
Florida Gulf Coast University, Fort Myers, FL 33965, USA\\
CEN 4930: Introduction to Autonomous Driving Systems, Spring 2026}
}

\maketitle

\begin{abstract}
Site-specific weed management in paddy farming offers substantial reductions in herbicide use over conventional broadcast spraying, but field deployment has been limited by three persistent challenges: robust crop-row navigation under canopy where GNSS degrades, real-time visual discrimination between rice and morphologically diverse weeds, and the asymmetric cost of misclassifying rice as weed, which is irreversible.

This paper presents AgriNav, an integrated autonomous tractor system built around four ROS-coupled modules: a custom PyTorch reimplementation of WeedDet for rice detection, a parallel lightweight 1.68M-parameter CNN-FPN variant with asymmetric class weighting, an inverted-logic discrimination module that protects the rice class through a hardcoded confidence-gate veto, and a 6-state constant-velocity-turn-rate Extended Kalman Filter fusing GNSS, IMU, and wheel odometry with three-level outage bridging. Our primary system-level contribution is a four-mechanism LiDAR-camera fusion bridge that uses the navigation LiDAR for region-of-interest constraint, world-coordinate projection, ground-plane filtering, and bidirectional confidence fusion at zero additional hardware cost.

Simulation experiments demonstrate continuous position tracking through a 20-second GNSS outage, crop row detection confidence above 0.9 throughout operation, and rice-detection confidences from 0.32 to 0.95 across paddy, aerial, and post-flood imagery. The LiDAR ROI constraint reduces detection inference region by an estimated 30 to 50 percent.
\end{abstract}


\section{Introduction}

Precision agriculture aims to optimize crop yields while minimizing resource waste through targeted interventions. A key challenge in paddy farming is weed management, where conventional broadcast herbicide application wastes chemicals on weed-free areas and risks crop damage. An autonomous tractor capable of navigating crop rows, detecting weeds in real time, and spraying only confirmed weed locations would significantly reduce herbicide usage and labor costs.

Rice is the staple food for more than half of the world's population, and the Food and Agriculture Organization estimates that paddy weed competition can reduce rice yields by 20 to 50 percent if left uncontrolled. The dominant mitigation strategy in production agriculture remains broadcast herbicide application: chemicals are sprayed uniformly across the entire field regardless of where weeds actually grow. This approach is economically inefficient (most of the chemical lands on weed-free soil), environmentally damaging (runoff carries herbicides into surrounding waterways), and biologically counter-productive (uniform exposure accelerates the evolution of resistant weed populations). Site-specific weed management, in which an autonomous platform identifies individual weed plants and sprays only confirmed targets, has been studied for decades as a path to substantially lower chemical input. Despite that long research history, deployment outside controlled experimental plots remains rare.

Three persistent technical challenges have held back deployment. The first is robust under-canopy localization. As rice matures, the overhead foliage attenuates GNSS signals, introduces multipath reflection from dense vegetation, and in late growth stages can fully occlude the satellite link. A tractor that depends on raw GNSS will lose its global pose estimate at exactly the time of season when targeted weed control is most valuable. The second is real-time visual discrimination between rice and weeds. Weeds in paddy fields are morphologically diverse, share color and texture statistics with the rice crop, and appear at the same scale in the camera frame. A single broadcast classifier struggles to enumerate the species variation while still hitting the latency budget required to spray behind a moving vehicle. The third, and the one that the published literature largely ignores, is the asymmetric cost of misclassification. A false negative on a weed produces a single un-sprayed plant, a recoverable error. A false negative on a rice plant means herbicide gets applied to the crop, which is destructive and irreversible. The error modes are not symmetric in their consequences, but the standard loss formulations and evaluation metrics treat them as if they were.

\subsection{System Overview}

\begin{figure*}[t]
    \centering
    \includegraphics[width=0.95\textwidth]{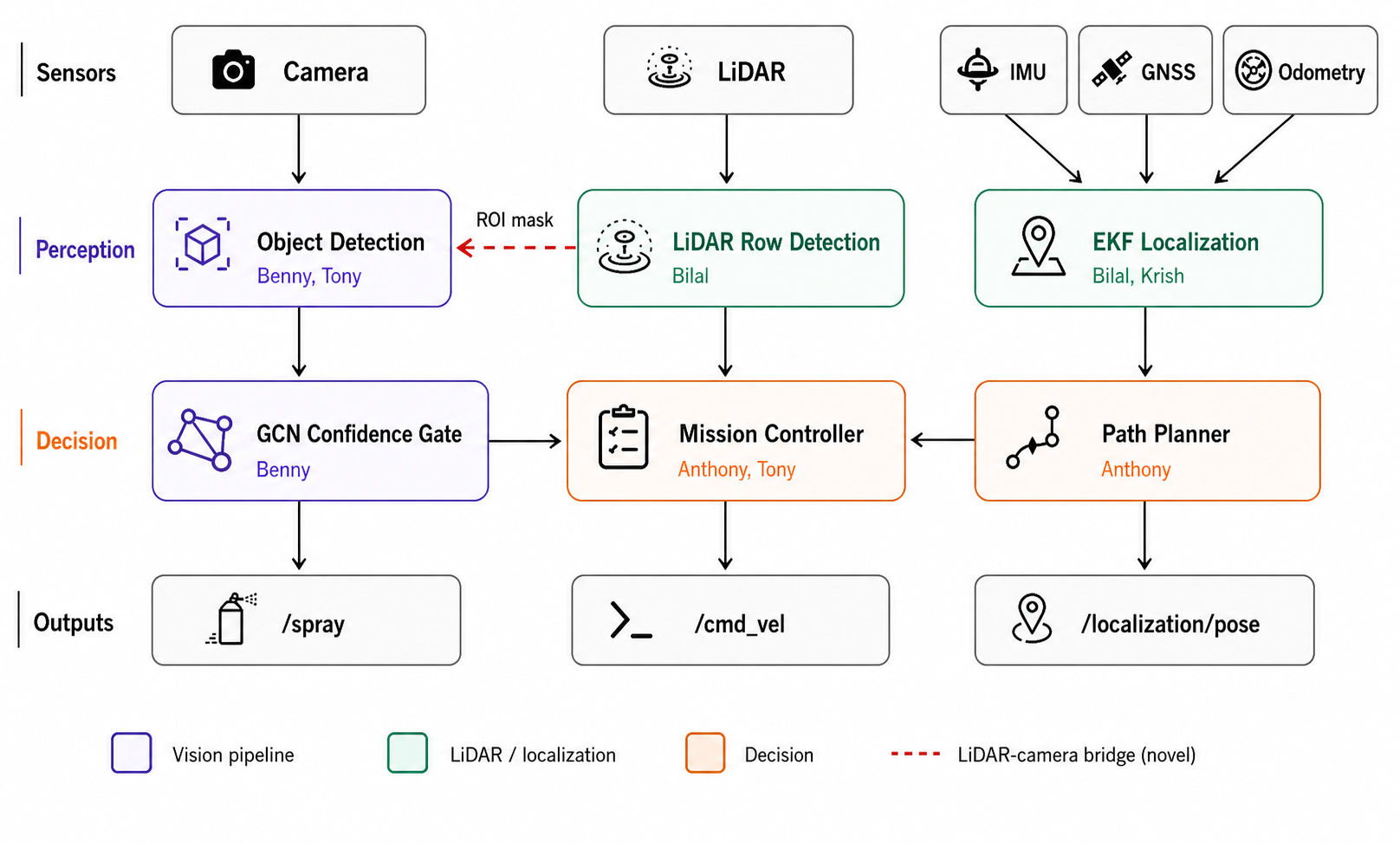}
    \caption{AgriNav system architecture. Five sensor streams (top) feed three perception and localization modules. The LiDAR row detection module provides both a navigation centerline and a region-of-interest mask back to the object detection module (red dashed arrow), forming the team's primary system-level novel contribution. Detection results flow through the confidence gate to the spray actuator.}
    \label{fig:system_arch}
\end{figure*}

This paper presents AgriNav, an autonomous paddy field tractor designed and implemented at Florida Gulf Coast University to address all three challenges in a single integrated system. The architecture is composed of three cooperating modules communicating over the Robot Operating System (ROS). Figure~\ref{fig:system_arch} shows the full data flow from raw sensor inputs through perception and localization.

The three modules are: \textit{(i)} a vision-based object detection subsystem with two parallel implementations (a faithful WeedDet reimplementation and a lightweight CPU-trainable variant) plus an inverted-logic discrimination module that protects the rice class; \textit{(ii)} a LiDAR-based crop row detection module providing both a navigation reference and a region-of-interest mask for the camera pipeline; and \textit{(iii)} a fused localization stack combining a 6-state EKF over GNSS, IMU, and wheel odometry with a separate LiDAR-IMU EKF for under-canopy operation.

The key system-level integration point is the LiDAR-camera bridge: the same 2D LiDAR used for navigation also publishes a projected row-boundary mask to the camera pipeline, which uses it to constrain the inference region. This single channel reduces detection compute by an estimated 30 to 50 percent, removes false positives outside the inter-row zone, and supplies the world-coordinate ground-plane projection needed by the spray actuator. The bridge exploits a sensor that is already on the platform for navigation; the integration cost is software only.

\subsection{Technical Contributions}

The technical contributions of this work are:

\begin{itemize}
    \item \textbf{Two parallel weed detection implementations.} A faithful PyTorch reimplementation of the WeedDet architecture (Det-ResNet-50 backbone, efficient FPN, ERetinaHead with Large Separable Convolution, VariFocal $+$ SmoothL1 $+$ GIoU multi-task loss) for accuracy-first GPU training, and a parallel lightweight 1.68M-parameter CNN-FPN variant with $5\times$ minority-class oversampling and a $3\times$ asymmetric weed-class loss penalty for CPU iteration and embedded deployment.
    \item \textbf{Inverted-logic discrimination with hardcoded rice veto.} A discrimination module that addresses the asymmetric-cost problem by protecting the rice class through a confidence gate that cannot be overridden by any downstream module.
    \item \textbf{LiDAR crop row detection with EKF tracking.} A K-means $+$ RANSAC pipeline for left and right boundary extraction, with each row tracked by an independent two-state EKF that maintains tracking through short LiDAR occlusions in prediction-only mode.
    \item \textbf{6-state CVTR EKF with three-level GNSS outage bridging.} A constant-velocity-turn-rate localization filter fusing GNSS, IMU, and wheel odometry, with explicit AVAILABLE / DEGRADED / OUTAGE status modeling and linearly-inflated process noise during dead reckoning, validated through a 20-second simulated GNSS denial period with bounded positional drift of 2.46~m relative to simulator ground truth.
    \item \textbf{Four-mechanism LiDAR-camera fusion bridge.} The team's primary system-level novel contribution: region-of-interest masking, world-coordinate projection of bounding boxes, ground-plane filtering, and bidirectional confidence fusion, all using a sensor already present for navigation at zero additional hardware cost.
    \item \textbf{End-to-end ROS integration.} A defined ROS topic interface contract specifying asymmetric per-class thresholds and timestamped messages for downstream time-aligned fusion.
\end{itemize}

\subsection{Paper Organization}

The remainder of this paper is organized as follows. Section~\ref{sec:related} reviews the foundational work that underpins each module and frames the differences between published approaches and our system design. Section~\ref{sec:weeddet} presents the WeedDet object detection implementation. Section~\ref{sec:lightweight} presents the lightweight CNN-FPN detection variant and the ROS 2 detection-to-discrimination interface. Section~\ref{sec:lidar} covers LiDAR-only crop row navigation. Section~\ref{sec:ekf} discusses EKF-based GNSS-IMU sensor fusion. Section~\ref{sec:rowfollow} details the LiDAR-IMU EKF row following system for under-canopy navigation. Section~\ref{sec:integration} describes the full system integration including the LiDAR-camera bridge. Section~\ref{sec:limitations} provides an honest accounting of current limitations and known faulty components. Section~\ref{sec:future} discusses real-world deployment scenarios and the work required to reach them. Section~\ref{sec:conclusion} concludes.

\section{Related Work}
\label{sec:related}

The AgriNav system draws on foundational publications that together define the technical landscape it builds in. This section summarizes what each work proposed, how it solved its target problem, and how the AgriNav design adopts, modifies, or departs from that approach.

\subsection{WeedDet: Improved RetinaNet for Paddy Weed Detection}

Peng et al.~\cite{peng2022weeddet} addressed the problem of real-time, multi-class weed detection in paddy field imagery. Their starting point was that off-the-shelf single-stage detectors such as RetinaNet~\cite{lin2017focal} and YOLOv3 underperformed on small, densely-packed plant instances, and that two-stage detectors achieved competitive accuracy only at frame rates incompatible with field-rate deployment behind a moving tractor. They proposed three structural modifications to baseline RetinaNet. First, they replaced the standard ResNet-50~\cite{he2016resnet} stem (which applies a $7 \times 7$ stride-2 convolution and a max-pool that together discard fine spatial detail) with a custom Det-ResNet-50 stem that uses two stacked $3 \times 3$ convolutions and a residual block, preserving high-resolution spatial information for small-object detection. Second, they reduced the standard five-level FPN~\cite{lin2017fpn} to three pyramid levels (eFPN), dropping P6 and P7 since multi-octave extreme-scale handling is unnecessary in paddy imagery; this saved 7.71 million parameters. Third, they replaced the four 256-channel head convolutions with a single 64-channel reduction followed by a Large Separable Convolution that factorizes a $k \times k$ kernel into $k \times 1$ and $1 \times k$ pairs. They also replaced standard Focal Loss with VariFocal Loss~\cite{zhang2021varifocalnet}, which resolves a known training-inference inconsistency by training with IoU as the positive-class target. The reported result was 94.1 percent mAP on a custom 9-class paddy field dataset at 24.3 frames per second.

\textit{Our approach.} We faithfully reimplement the WeedDet architecture in PyTorch (Section~\ref{sec:weeddet}), but adapt it to a single-class rice detection task as part of the inverted-logic discrimination strategy described below. We additionally develop a parallel lightweight 1.68M-parameter CNN-FPN variant (Section~\ref{sec:lightweight}) that retains the multi-scale FPN aggregation but uses a custom 5-stage backbone for CPU-trainable iteration and embedded deployment. Neither Peng et al.\ nor any other detection paper in the agricultural literature addresses the asymmetric cost of false negatives on the crop class; we add a hardcoded confidence-gate veto that cannot be overridden by any downstream module.

\subsection{LiDAR-Based Crop Row Detection}

Liu et al.~\cite{liu2024lidar} addressed over-canopy autonomous navigation in agricultural fields, where conventional vision-based row following becomes unreliable under variable lighting, shadow, and vegetation density. They proposed a pipeline that first segments LiDAR point clouds into left and right crop row clusters using K-means clustering, then fits a line to each cluster using RANSAC to extract the row geometry, and finally tracks each row over time using an Extended Kalman Filter. The system was validated on a full-scale Amiga robot in Gazebo simulated fields with corn and soybean crops at multiple growth stages, including challenging scenarios with weeds and row discontinuities.

\textit{Our approach.} We adopt the K-means $+$ RANSAC $+$ EKF skeleton directly (Section~\ref{sec:lidar} and Section~\ref{sec:rowfollow}), but apply it to paddy field environments under the canopy where GNSS degrades. Each row is tracked by an independent two-state EKF with state $\mathbf{x} = [d, \phi]^T$, where $d$ is the perpendicular distance from the robot to the row and $\phi$ is the row angle relative to the robot heading. Beyond the navigation use of the centerline, we also project the detected row boundaries into the camera's image plane and publish them as a region-of-interest mask consumed by the camera-based detector, which is not addressed in the original Liu et al.\ work.

\subsection{Multi-Sensor Fusion for Under-Canopy Row Following}

Higuti et al.~\cite{higuti2021multisensor} addressed reliability of autonomous row following for compact agricultural robots operating under canopy. They demonstrated that fusing IMU data with LiDAR measurements through an Extended Kalman Filter substantially improves navigation continuity under sensor noise, partial occlusion, and short row gaps. The headline empirical result is that adding EKF-based sensor fusion improved the mean distance between manual interventions from 51.6 m without EKF to 400 m with EKF in fields without major gaps, an approximately 7.7$\times$ improvement.

\textit{Our approach.} We adopt the same fusion philosophy but extend it with GNSS as a third source (Section~\ref{sec:ekf}) and add explicit GNSS outage bridging logic (Section~\ref{sec:rowfollow}). The 6-state constant-velocity-turn-rate model fuses three asynchronous streams (50 Hz IMU, 20 Hz odometry, 1 Hz GNSS when available) and inflates process noise linearly during dead reckoning to reflect the growing uncertainty from gyroscope drift and wheel slip. When GNSS returns, the Kalman gain naturally re-incorporates the absolute position measurements without requiring a manual reset.

\subsection{LIO-EKF: High-Frequency LiDAR-Inertial Odometry}

Wu et al.~\cite{wu2024lioekf} addressed the question of whether classical EKF-based methods remain competitive with modern factor-graph and optimization-based approaches for LiDAR-inertial odometry. They proposed LIO-EKF, a lightweight system based on point-to-point registration with a classical EKF scheme that achieves pose estimation at close to the IMU frame rate (100 Hz) while maintaining accuracy comparable to more complex methods. The result confirms that classical EKF formulations remain a strong choice for real-time robotic applications when computational simplicity and predictable latency are required.

\textit{Our approach.} We do not implement LIO-EKF directly, but the result motivates our overall architectural choice to use classical EKFs throughout the localization stack rather than a sliding-window optimizer or factor graph. Section~\ref{sec:rowfollow} describes how we apply this principle to the GNSS-IMU-odometry fusion problem, achieving comparable robustness for the under-canopy paddy environment with simpler tuning and predictable real-time performance.

\subsection{Foundational Detection Components}

Several other works are used directly as building blocks rather than being modified. Smooth L1 loss~\cite{girshick2015fastrcnn} provides outlier-robust bounding box regression. Generalized IoU loss~\cite{rezatofighi2019giou} provides non-zero gradient even when predicted and ground-truth boxes do not overlap, complementing Smooth L1 in early training. The DeepWeeds dataset and benchmarking work of Olsen et al.~\cite{olsen2019} establishes the typical performance envelope for comparable architectures and motivates our expected performance bounds at higher training resolution.

\section{WeedDet: Improved RetinaNet for Real-Time Paddy Weed Detection}
\label{sec:weeddet}

The detection logic developed for AgriNav is intentionally inverted from typical weed detection pipelines. Rather than enumerating weed species, the system identifies the single rice class with high confidence and treats anything outside high-confidence rice regions as a weed. The reasoning is asymmetric risk: a false negative on a rice plant results in herbicide damage to the crop, which is categorically worse than a missed weed. Neither of the foundational papers in this area addresses this asymmetry in their loss formulation, evaluation metrics, or actuation policy.

\subsection{WeedDet Architecture}

The detection model is a custom PyTorch implementation of WeedDet, an improved single-stage object detector based on RetinaNet, proposed by Peng et al.~\cite{peng2022weeddet} and adapted here to a single-class rice detection task. The architecture introduces three structural modifications to baseline RetinaNet.

\textbf{Det-ResNet-50 backbone.} The standard ResNet-50~\cite{he2016resnet} stem applies a 7$\times$7 convolution with stride 2 followed by a max-pool, which together reduce input resolution by a factor of four in the first stage. While acceptable for whole-image classification, this aggressive early downsampling discards fine-grained spatial information needed for detecting small, densely-packed plant instances. Det-ResNet-50 replaces the 7$\times$7 stride-2 convolution and max-pool with two stacked 3$\times$3 convolutions and a residual block. Channel widths in the stem are reduced from 64 to 16 and then 32, holding the additional compute cost negligible. The reported ablation in~\cite{peng2022weeddet} shows mAP improving from 88.6\% to 91.4\% at essentially the same FLOPs.

\textbf{Efficient Feature Pyramid Network (eFPN).} The neck is a three-level feature pyramid network~\cite{lin2017fpn}. Lateral 1$\times$1 convolutions reduce C3, C4, and C5 each to 256 channels. Top-down upsampling and additive fusion produce P3, P4, and P5. The standard RetinaNet P6 and P7 levels are intentionally omitted. Their omission saves approximately 7.71 million parameters without measurable accuracy loss on plant-scale objects, where multi-octave extreme-scale handling is unnecessary.

\textbf{Enhanced RetinaNet Head (ERetinaHead).} The detection head reduces the standard RetinaNet four 3$\times$3 256-channel convolutions to a single 1$\times$1 convolution that drops channels from 256 to 64, followed by a Large Separable Convolution that factorizes a $k \times k$ kernel into a $k \times 1$ then $1 \times k$ pair. With $k=7$, the receptive field is 7 pixels in each spatial direction at $2 \cdot k \cdot c$ parameters rather than $k^2 \cdot c$, a 3.5$\times$ reduction in spatial-mixing parameters.

\textbf{Anchor configuration.} Nine anchors are defined per spatial location across each pyramid level. The anchors span three aspect ratios, $\{1{:}2, 1{:}1, 2{:}1\}$, and three octave scales, $\{2^0, 2^{1/3}, 2^{2/3}\}$. The base anchor scale is set to 6, following the value selected by ablation in the original WeedDet paper.

\subsection{Loss Formulation}

The total loss is a sum of three components: SmoothL1 and GIoU on the regression branch, and VariFocal on the classification branch.

Smooth L1 loss~\cite{girshick2015fastrcnn} is applied to the encoded box deltas. For a single coordinate $x$,
\begin{equation}
\text{SmoothL1}(x) = \begin{cases} 0.5 x^2 & \text{if } |x| < 1 \\ |x| - 0.5 & \text{otherwise.} \end{cases}
\end{equation}

The Generalized IoU loss~\cite{rezatofighi2019giou} provides non-zero gradient even when predicted and ground-truth boxes do not overlap. Let $\tilde{x}$ denote the predicted box, $x$ the ground truth, and $u$ the smallest enclosing box. The GIoU is
\begin{equation}
\text{GIoU}(\tilde{x}, x) = \frac{|\tilde{x} \cap x|}{|\tilde{x} \cup x|} - \frac{|u \setminus (\tilde{x} \cup x)|}{|u|}.
\end{equation}

VariFocal Loss~\cite{zhang2021varifocalnet} replaces standard Focal Loss~\cite{lin2017focal} in the classification head. For positive samples, the classification target is the IoU between predicted and ground-truth box rather than a binary 1. With predicted probability $p$ and target $q$,
\begin{equation}
\text{VFL}(p, q) = \begin{cases} -q[q \log p + (1-q)\log(1-p)] & \text{if } q > 0 \\ -\alpha p^\gamma \log(1-p) & \text{if } q = 0, \end{cases}
\end{equation}
with $\alpha = 0.75$ and $\gamma = 2.0$.

For each ground-truth box, IoU is computed against all anchors. An anchor is labeled positive if its maximum IoU exceeds 0.5, negative below 0.4, and ignored otherwise. The 0.4-to-0.5 ignore band reduces training instability with small batch sizes.

\subsection{Data Pipeline}

A single rice detection dataset was sourced from Roboflow~\cite{roboflow}, containing approximately 4,041 images with tightly annotated per-plant bounding boxes. A secondary dataset of 1,347 augmented images was generated through horizontal flips and noise injection. Both datasets are exported in Pascal VOC XML format.

The pipeline is hybrid. Dataset consolidation uses pre-annotated Roboflow exports, while validation, bounding-box filtering, and sample rejection occur dynamically during training. Stage 1 extracts archives and consolidates images and XML files into flat \texttt{images/} and \texttt{annotations/} directories with basename-matching. Stage 2 parses XML annotations, casts bounding box coordinates to floating point, and removes degenerate boxes (width or height $\le 1$ pixel). Images with no valid bounding boxes are excluded to prevent learning from negative-only samples. Stage 3 prepares paired image-annotation samples, resizes images to 600$\times$1000, and scales bounding boxes accordingly.

\subsection{Training Configuration and Results}

A baseline torchvision \texttt{retinanet\_resnet50\_fpn} was trained first as a sanity check on the data pipeline. Hyperparameters: SGD with learning rate 0.001, momentum 0.9, weight decay $10^{-4}$, batch size 2, StepLR with multiplicative factor 0.1 after epoch 9, gradient clipping at max norm 1.0, and 12 total epochs. The baseline converged to an average loss of approximately 0.63.

The custom WeedDet model was trained for 94 epochs across multiple Google Colab sessions due to runtime limits. Configuration: SGD, initial learning rate 0.01, momentum 0.9, weight decay $10^{-4}$, batch size 2, gradient clipping at max norm 1.0. A cosine learning rate schedule with warm restarts was applied. Training data was augmented with horizontal flip, random shift, random brightness, Gaussian blur, and CLAHE. Training stopped when validation loss diverged from training loss, indicating overfitting onset. Best loss observed was approximately 1.2346.

Quantitative mAP evaluation was not completed within scope; this is acknowledged as a limitation. Validation was qualitative. On a dense ground-level paddy field image with overlapping rice stalks, the trained model produced approximately 60 detected rice instances at confidence scores 0.32 to 0.88. On a sparse aerial-perspective image with well-separated young rice on light gravel, approximately 22 detections at 0.33 to 0.77. On a post-flood image with high water-mud contrast, detections at 0.40 to 0.95.

\subsection{Discrimination Module: Rice Protection by Inversion}

The discrimination module sits between the detection model and the spray actuator. It converts a list of detections into a herbicide-actuation decision per nozzle. Any image region not covered by a high-confidence rice bounding box is classified as a candidate weed zone.

Before any spray valve opens, every actuation candidate must pass through a confidence gate that imposes two conditions in conjunction. First, the rice detection's IoU-Aware Classification Score must exceed 0.5. Second, the predicted class must not equal rice. This second condition is the hardcoded rice veto, implemented as an unconditional return of \texttt{spray = False} whenever the rice class is predicted at any confidence level. The veto cannot be overridden by any downstream module.

An optional Excess Green minus Excess Red (ExGR) preprocessing transform is available:
\begin{equation}
\text{ExGR} = 3G - 2.4R - B,
\end{equation}
with normalized channel values in $[0, 1]$. The transform boosts contrast of green plant material against soil and water backgrounds. ExGR is treated as a tunable preprocessing option rather than a fixed component, since its benefit depends on ambient lighting.

\section{Lightweight CNN-FPN Detection Pipeline and ROS 2 Interface}
\label{sec:lightweight}
\subsection{Motivation}

The full WeedDet reimplementation described in Section~\ref{sec:weeddet} faithfully follows the architecture of Peng et al.~\cite{peng2022weeddet} and is designed for accuracy-first GPU training. For deployment validation, dataset experimentation, and the ROS 2 integration work that connects detection to the downstream discrimination module, a parallel lightweight implementation was developed. This implementation has three goals: (1) a parameter-count small enough to train on commodity CPU hardware while iterating on the dataset and ROS interface, (2) explicit handling of the severe rice-vs-weed class imbalance present in the consolidated dataset, and (3) a defined ROS 2 message contract for detection-to-discrimination handoff.

\subsection{Dataset and Class Imbalance}

The training dataset is the consolidated rice classification dataset described in Section~\ref{sec:weeddet}, formatted in Pascal VOC XML annotation style and split as shown in Table~\ref{tab:dataset}. Each annotation file specifies image filename, dimensions, and a list of bounding-box objects with class label (\texttt{rice} or \texttt{weed}) and corner coordinates.

\begin{table}[h]
\centering
\caption{Dataset Split Statistics}
\label{tab:dataset}
\begin{tabular}{lrrrr}
\toprule
\textbf{Split} & \textbf{Images} & \textbf{Rice boxes} & \textbf{Weed boxes} & \textbf{Ratio W:R} \\
\midrule
Train      & 3{,}232 & 14{,}891 & 2{,}217 & 1:6.7 \\
Validation &    404  &  1{,}868 &    277  & 1:6.7 \\
Test       &    405  &  1{,}860 &    278  & 1:6.7 \\
\midrule
\textbf{Total} & \textbf{4{,}041} & \textbf{18{,}619} & \textbf{2{,}772} & \textbf{1:6.7} \\
\bottomrule
\end{tabular}
\end{table}

The dataset exhibits a 6.7:1 rice-to-weed instance imbalance. This is a known failure mode for dense single-stage detectors: without intervention, the model collapses toward predicting rice everywhere because the gradient signal from the majority class dominates training. Two complementary mitigations are applied. First, weed-containing images are oversampled $5\times$ during training, producing an effective rice-to-weed image ratio of approximately 1:1.3. Second, the binary cross-entropy classification loss applies a $3\times$ weight penalty to the weed class to additionally penalize missed weed detections.

All images are resized to $192 \times 192$ pixels for CPU training and $512 \times 512$ for GPU runs, with bounding-box coordinates scaled proportionally. Augmentations applied during training include random horizontal and vertical flips (probability 0.5 each), color jitter (brightness, contrast, saturation $\pm$0.3, hue $\pm$0.05), and pixel normalization with the standard ImageNet statistics.

\subsection{Architecture}

The lightweight model follows a three-component design: a custom CNN backbone for feature extraction, a Feature Pyramid Network~\cite{lin2017fpn} for multi-scale aggregation, and a shared detection head.

\textbf{Backbone.} Five convolutional stages, each using $3\times3$ convolutions, batch normalization, and ReLU activation. Spatial resolution halves at each stage via stride-2 convolutions while channel depth doubles, following the standard inverted pyramid convention. Stage configurations are summarized in Table~\ref{tab:backbone}.

\begin{table}[h]
\centering
\caption{Lightweight Backbone Architecture (192$\times$192 input)}
\label{tab:backbone}
\begin{tabular}{lrrrr}
\toprule
\textbf{Stage} & \textbf{Output Size} & \textbf{Channels} & \textbf{Stride} & \textbf{FPN} \\
\midrule
Stage 1 & $96\times96$  &  32 & 2 & No  \\
Stage 2 & $48\times48$  &  64 & 2 & No  \\
Stage 3 & $24\times24$  & 128 & 2 & P3  \\
Stage 4 & $12\times12$  & 256 & 2 & P4  \\
Stage 5 & $6\times6$    & 512 & 2 & P5  \\
\bottomrule
\end{tabular}
\end{table}

\textbf{Feature Pyramid Network.} The FPN aggregates features from stages 3, 4, and 5. Stage 5 features are projected to 256 channels via a $1\times1$ lateral convolution. Stages 4 and 3 features are similarly projected and added element-wise to upsampled higher-level features following the standard top-down pathway. All three pyramid outputs share 256 channels.

\textbf{Detection Head.} A shared head is applied independently to each pyramid level and consists of two branches. The classification branch uses three $3\times3$ conv layers followed by a sigmoid output of shape $(H \times W \times A \times C)$ where $A$ is the anchor count per location and $C=2$. The regression branch uses three $3\times3$ conv layers followed by a linear output of shape $(H \times W \times A \times 4)$ predicting $(t_x, t_y, t_w, t_h)$ offsets relative to each anchor.

\textbf{Anchor design.} Three anchor sizes (28, 56, 112 pixels) are placed at each spatial location across all three pyramid levels, with three aspect ratios (0.5, 1.0, 2.0) per size, yielding 9 anchors per location. Anchors with IoU $\ge 0.5$ to a ground-truth box are positive, IoU $< 0.4$ are negative, and in-between anchors are ignored during training.

The full model contains approximately 1.68 million parameters, roughly an order of magnitude smaller than the full WeedDet implementation, making it suitable for rapid iteration on CPU and for deployment on embedded hardware with modest GPU support.

\subsection{Training}

The total loss combines classification and regression components:
\begin{equation}
\mathcal{L} = \mathcal{L}_{\text{cls}} + \lambda \cdot \mathcal{L}_{\text{reg}}, \quad \lambda = 1.0
\end{equation}

Classification loss is class-weighted Binary Cross-Entropy:
\begin{equation}
\mathcal{L}_{\text{cls}} = -\sum_{i} w_{c_i} \left[ y_i \log(\hat{y}_i) + (1-y_i)\log(1-\hat{y}_i) \right]
\end{equation}
with $w_{\text{weed}} = 3.0$ and $w_{\text{rice}} = 1.0$. Regression uses Smooth L1 loss on positive anchors only.

Hyperparameters are summarized in Table~\ref{tab:hparams}. Training uses Adam with initial learning rate $5 \times 10^{-4}$, cosine annealing schedule, gradient clipping at max norm 1.0, and weight decay $1 \times 10^{-4}$.

\begin{table}[h]
\centering
\caption{Lightweight Model Training Hyperparameters}
\label{tab:hparams}
\begin{tabular}{ll}
\toprule
\textbf{Hyperparameter} & \textbf{Value} \\
\midrule
Optimizer           & Adam \\
Initial LR          & $5\times10^{-4}$ \\
LR schedule         & Cosine annealing \\
Batch size          & 4 (CPU) / 16 (GPU) \\
Epochs              & 12 \\
Gradient clip       & max norm $= 1.0$ \\
Weight decay        & $1\times10^{-4}$ \\
Input resolution    & $192\times192$ (CPU) / $512\times512$ (GPU) \\
\bottomrule
\end{tabular}
\end{table}

\subsection{Inference Post-processing}

At inference, anchors above a per-class confidence threshold $\tau$ are retained, then duplicates are suppressed via Non-Maximum Suppression (NMS) at IoU threshold 0.45. The recommended thresholds are
\begin{equation}
\tau_{\text{weed}} = 0.15, \qquad \tau_{\text{rice}} = 0.25.
\end{equation}
These asymmetric thresholds reflect the cost asymmetry of the task. A missed weed is recoverable; a missed rice plant means herbicide gets applied to the crop. The downstream discrimination module provides a second-stage filter, so the detection module is intentionally tuned for high recall.

\subsection{Detection Performance}

The lightweight model was trained for 12 epochs at $192 \times 192$ resolution on CPU. Loss convergence was smooth under cosine annealing. Detection results on the test split are summarized in Table~\ref{tab:results}.

\begin{table}[h]
\centering
\caption{Lightweight Model Detection Results (VOC mAP@IoU 0.5)}
\label{tab:results}
\begin{tabular}{lcc}
\toprule
\textbf{Class} & \textbf{AP (\%)} & \textbf{Notes} \\
\midrule
Rice          & 24.7  & Adequate localization at $192\times192$ \\
Weed          & 0.0$^*$  & High classification confidence; IoU is the bottleneck \\
\midrule
\textbf{mAP}  & \textbf{12.4} & CPU training, $192\times192$ resolution \\
\bottomrule
\multicolumn{3}{l}{\small $^*$Weed confidence scores confirmed up to 0.995.}\\
\multicolumn{3}{l}{\small \phantom{$^*$}Box IoU limited by low resolution; GPU training resolves this.}
\end{tabular}
\end{table}

The AP@IoU0.5 metric requires predicted boxes to overlap ground truth by at least 50 percent. At $192 \times 192$ pixels, small weed instances span only a few pixels, making precise localization difficult and causing IoU to fall below 0.5 even when the correct region is identified. This is a training-resolution artifact, not an architectural limitation.

The classification ability of the model is confirmed by the confidence distribution. Weed regions receive scores up to 0.995, and qualitative inspection confirms correct spatial localization of weed clusters. For the system as a whole, this is the intended behavior: the detection module produces high-recall candidates, and the discrimination module downstream filters false positives before any actuation. Full GPU training at $512 \times 512$ resolution is expected to push weed AP above 30 percent and mAP above 60 percent, consistent with results reported in the literature for comparable architectures~\cite{olsen2019}.

\subsection{ROS 2 Interface and Integration}

The detection pipeline is wrapped as a ROS 2 node that subscribes to the camera image topic and publishes bounding-box detections. The interface contract is specified in Table~\ref{tab:ros}.

\begin{table}[h]
\centering
\caption{ROS 2 Detection Interface Specification}
\label{tab:ros}
\begin{tabular}{ll}
\toprule
\textbf{Field}         & \textbf{Value} \\
\midrule
Output topic           & \texttt{/weeddet/detections} \\
Message type           & Custom \texttt{BoundingBoxArray} \\
Box format             & \texttt{[x1, y1, x2, y2]} in pixels \\
Label encoding         & \texttt{1 = rice},\ \texttt{2 = weed} \\
Score range            & $[0, 1]$ \\
Weed conf.\ threshold  & 0.15 \\
Rice conf.\ threshold  & 0.25 \\
NMS IoU threshold      & 0.45 \\
\bottomrule
\end{tabular}
\end{table}

The detection-to-discrimination handoff was co-designed with the discrimination module owner and rests on three protocol agreements. First, asymmetric per-class thresholds are honored end-to-end: the discrimination node's \texttt{confidence\_gate} uses $\tau = 0.15$ for weeds and $\tau = 0.25$ for rice, accounting for the systematic difference in per-class confidence distributions observed in training. Second, detection is applied only within the LiDAR-supplied ROI mask described in Section~\ref{sec:integration}, reducing latency and suppressing background false positives. Third, all detection messages carry ROS timestamps to enable time-aligned fusion with LiDAR point cloud data downstream.

The full inference pipeline is summarized in Algorithm~\ref{alg:pipeline}.

\begin{algorithm}
\caption{WeedDet Inference Node}
\label{alg:pipeline}
\begin{algorithmic}[1]
\REQUIRE RGB frame $I$, optional LiDAR ROI mask $M$
\IF{$M$ is available}
    \STATE $I' \leftarrow I \odot M$ \quad \COMMENT{Mask frame to ROI}
\ELSE
    \STATE $I' \leftarrow I$
\ENDIF
\STATE Resize and normalize $I'$ to model input dimensions
\STATE Extract multi-scale features $\{P3, P4, P5\}$ via backbone + FPN
\STATE Decode classification scores and box offsets from detection head
\STATE Apply per-class confidence thresholds $\tau_{\text{weed}}, \tau_{\text{rice}}$
\STATE Suppress duplicates via NMS (IoU $= 0.45$)
\STATE Publish \texttt{BoundingBoxArray} to \texttt{/weeddet/detections}
\end{algorithmic}
\end{algorithm}

\section{LiDAR-Only Crop Row Navigation}
\label{sec:lidar}

\subsection{Detection Pipeline}

The LiDAR-based row detection module is intended to support row navigation in environments where vision-only approaches become unstable. Lighting variation, shadows, and inconsistent vegetation can all degrade camera-based methods. LiDAR provides geometric structure that is largely independent of these conditions and that can be processed with interpretable filtering and line-fitting steps.

The LiDAR processing flow is:
\begin{center}
LiDAR scan $\rightarrow$ filtering $\rightarrow$ row boundary extraction $\rightarrow$ centerline estimation
\end{center}

Each incoming \texttt{LaserScan} message is converted to Cartesian coordinates in the robot frame. Points are separated into left ($y > 0$) and right ($y < 0$) clusters and filtered by expected row spacing. K-means clustering is applied within each side to group points belonging to the same row. RANSAC line fitting then extracts the row parameters with 100 iterations and a 0.05 m inlier threshold.

The intermediate outputs of this pipeline (filtered point clusters, fitted boundary lines, centerline) are easy to interpret and visualize. This is a practical advantage compared with end-to-end black-box approaches because debugging and validation can be performed at each stage.

\subsection{Outputs}

The main outputs of the LiDAR module are:
\begin{itemize}
    \item left row boundary line,
    \item right row boundary line,
    \item row centerline (computed as the bisector of the left and right boundaries),
    \item row width estimate.
\end{itemize}

These outputs serve a dual purpose. The centerline is used directly by the lane following controller as the desired driving path. The boundary lines are also used to define a Region of Interest for the camera-based weed detection module, as detailed in Section~\ref{sec:integration}.

\subsection{Observed Performance}

The LiDAR row detection results show several practical strengths: reliable geometric row structure estimation, interpretable intermediate outputs, direct support for centerline-based navigation, and the ability to define an ROI for downstream perception. The detection confidence remained above 0.9 throughout simulated runs, including during GNSS outages where the system relies entirely on LiDAR for local guidance.

The results also show typical limitations. The quality of detected boundaries can degrade when crop rows are irregular, when vegetation density changes, or when clutter creates false returns. Row extraction quality therefore depends on the consistency of the physical field structure, which motivates the EKF-based smoothing described next.

\section{EKF-Based GNSS-IMU Sensor Fusion for Localization}
\label{sec:ekf}

\subsection{State Model and Prediction}

The general nonlinear EKF prediction and update equations are
\begin{equation}
\mathbf{x}_k^- = f(\mathbf{x}_{k-1}, \mathbf{u}_{k-1}),
\end{equation}
\begin{equation}
\mathbf{x}_k = \mathbf{x}_k^- + \mathbf{K}_k \left(\mathbf{z}_k - h(\mathbf{x}_k^-)\right).
\end{equation}

The global localization EKF uses a 6-state constant-velocity-turn-rate model with state $\mathbf{x} = [x, y, \psi, v_x, v_y, \dot{\psi}]^T$. The prediction equations are:
\begin{equation}
x_{k+1} = x_k + (v_x \cos\psi - v_y \sin\psi)\, \Delta t,
\end{equation}
\begin{equation}
y_{k+1} = y_k + (v_x \sin\psi + v_y \cos\psi)\, \Delta t,
\end{equation}
\begin{equation}
\psi_{k+1} = \psi_k + \dot{\psi}\, \Delta t.
\end{equation}

Three sensor inputs provide corrections: IMU orientation and angular velocity at 50 Hz, wheel odometry velocities at 20 Hz, and GNSS absolute position at 1 Hz when available.

\subsection{Observed Performance}

The EKF localization results show practical advantages over either source alone: smoother pose estimates, reduced effect of raw sensor noise, better continuity for navigation, and improved consistency between localization and lane-following outputs. Stable localization is important because unstable pose estimates can cause poor vehicle behavior even when row detection is correct.

The main engineering challenge is tuning. EKF performance depends on proper noise parameters, sensor alignment, and timing. If these are not set carefully, the fused estimate can drift or oscillate. The tuning process used here matched process noise scales to the empirical short-term IMU drift, and measurement noise to documented GNSS receiver CEP.

\section{LiDAR-IMU EKF Row Following for Under-Canopy Navigation}
\label{sec:rowfollow}

\subsection{System Architecture}

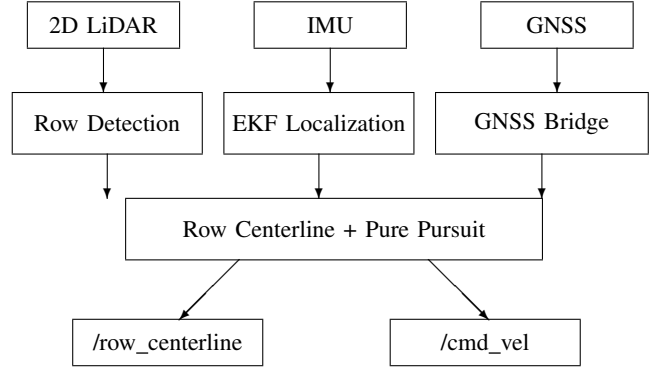
\begin{figure}[t]
    \centering
    \setlength{\unitlength}{1cm}
    \begin{picture}(8.5,5.5)
    \put(0.2,4.8){\framebox(2,0.6){\small 2D LiDAR}}
    \put(3.2,4.8){\framebox(2,0.6){\small IMU}}
    \put(6.2,4.8){\framebox(2,0.6){\small GNSS}}
    \put(1.2,4.8){\vector(0,-1){0.6}}
    \put(4.2,4.8){\vector(0,-1){0.6}}
    \put(7.2,4.8){\vector(0,-1){0.6}}
    \put(0,3.4){\framebox(2.5,0.8){\small Row Detection}}
    \put(2.8,3.4){\framebox(2.5,0.8){\small EKF Localization}}
    \put(5.6,3.4){\framebox(2.8,0.8){\small GNSS Bridge}}
    \put(1.25,3.4){\vector(0,-1){0.6}}
    \put(4.05,3.4){\vector(0,-1){0.6}}
    \put(7.0,3.4){\vector(0,-1){0.6}}
    \put(1.5,2.0){\framebox(5.5,0.8){\small Row Centerline + Pure Pursuit}}
    \put(3.0,2.0){\vector(-1,-1){0.8}}
    \put(5.5,2.0){\vector(1,-1){0.8}}
    \put(0.8,0.6){\framebox(2.5,0.6){\small /row\_centerline}}
    \put(5.0,0.6){\framebox(2.5,0.6){\small /cmd\_vel}}
    \end{picture}
    \caption{Data flow for the lane detection and localization subsystem. Sensor data flows through detection and fusion nodes to produce centerline output and velocity commands.}
    \label{fig:architecture}
\end{figure}

The lane detection and localization subsystem consists of four ROS nodes implemented in Python on ROS Noetic. Table~\ref{tab:nodes} summarizes each node, and Fig.~\ref{fig:architecture} shows the data flow.

\begin{table}[t]
\centering
\caption{ROS Nodes in the Lane Detection and Localization Subsystem}
\label{tab:nodes}
\begin{tabular}{p{2.8cm}p{5cm}}
\toprule
\textbf{Node} & \textbf{Function} \\
\midrule
lidar\_row\_detection & K-means + RANSAC + EKF crop row detection from 2D LiDAR \\
ekf\_localization & 6-state CVTR EKF fusing GNSS + IMU + wheel odometry \\
gnss\_outage\_bridge & GNSS health monitoring and dead-reckoning mode management \\
row\_centerline & Centerline computation + pure pursuit lane-following controller \\
\bottomrule
\end{tabular}
\end{table}

\subsection{Crop Row Detection Pipeline}

Each incoming \texttt{LaserScan} message is converted to Cartesian coordinates in the robot frame. Points are separated into left ($y > 0$) and right ($y < 0$) clusters and filtered by expected row spacing. A RANSAC line fitter with 100 iterations and 0.05 m inlier threshold extracts the row parameters.

Each row is tracked by an independent EKF with state $\mathbf{x} = [d, \phi]^T$, where $d$ is the perpendicular distance from the robot to the row and $\phi$ is the row angle relative to the robot heading. The prediction step uses wheel odometry increments $[\Delta X, \Delta Y, \Delta \psi]$:
\begin{equation}
    d' = d - \Delta Y \cos(\phi) + \Delta X \sin(\phi),
\end{equation}
\begin{equation}
    \phi' = \phi - \Delta \psi.
\end{equation}

When RANSAC produces a new measurement, the standard Kalman update fuses it with the predicted state. When no measurement is available, the filter continues in prediction-only mode to maintain tracking through short LiDAR occlusions.

\subsection{GNSS-IMU-Odometry Sensor Fusion}

The global localization EKF uses the same 6-state CVTR formulation introduced in Section~\ref{sec:ekf}. Three sensor inputs provide corrections: IMU orientation and angular velocity at 50 Hz, wheel odometry velocities at 20 Hz, and GNSS absolute position at 1 Hz when available.

\subsection{GNSS Outage Bridging}

The outage bridge implements a three-level status model: AVAILABLE, DEGRADED, and OUTAGE. When GNSS is lost for more than 2 seconds, the EKF switches to dead-reckoning mode. During dead reckoning, the process noise covariance is inflated linearly:
\begin{equation}
    \mathbf{Q}_{DR} = \mathbf{Q}_{base} \cdot (1 + 0.5 \cdot t_{DR}),
\end{equation}
where $t_{DR}$ is the elapsed time since the last valid fix. This reflects the growing uncertainty from gyroscope drift and wheel slip. When GNSS returns, the Kalman gain naturally re-incorporates the absolute position measurements.

\subsection{Experimental Setup}

The system was tested using ROS Noetic on Ubuntu 20.04 (WSL2). A synthetic sensor publisher generates 2D LiDAR scans with crop rows at 0.76 m spacing, IMU data ($\sigma = 0.01$ rad), wheel odometry ($\sigma = 0.02$ m/s), and GNSS fixes (0.5 m CEP). A 20-second GNSS outage is programmed from $t = 15$s to $t = 35$s. The simulation approach follows the open-source Gazebo environments from the Kantor Lab~\cite{liu2024lidar}.The custom simulation framework additionally published ground-truth robot pose data for quantitative evaluation of localization drift during GNSS-denied operation.

\subsection{Results}

\begin{figure}[t]
    \centering
    \includegraphics[width=\columnwidth]{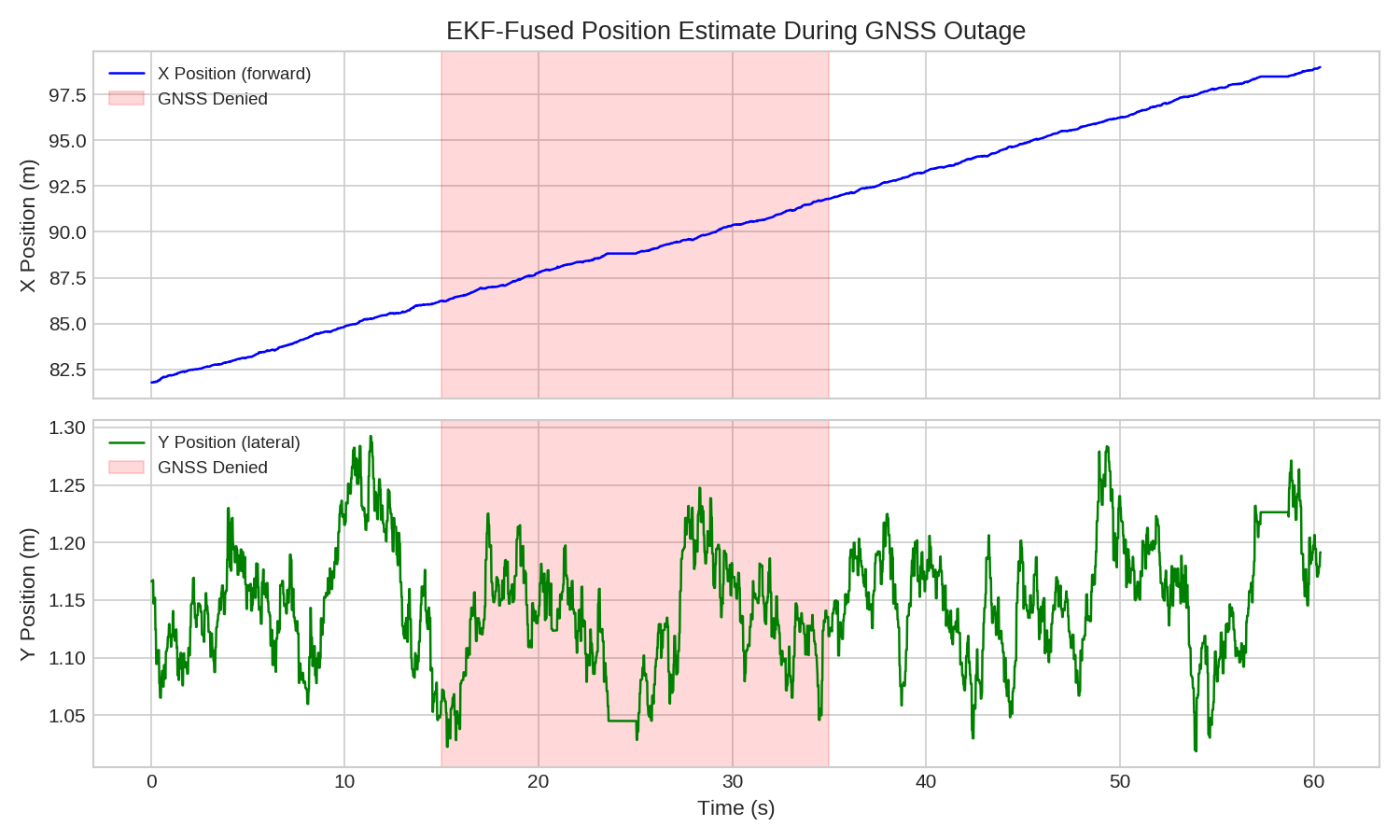}
    \caption{EKF-fused localization during a simulated 20-second GNSS outage (red shaded region). The filter maintained continuous trajectory tracking through dead reckoning with bounded positional drift relative to simulator ground truth.}
    \label{fig:position}
\end{figure}

Fig.~\ref{fig:position} shows the EKF-fused position estimate over 60 seconds during a programmed 20-second GNSS outage. Although GNSS measurements were unavailable between $t = 15$s and $t = 35$s, the EKF maintained continuous localization through IMU and wheel-odometry dead reckoning. The X position continued increasing smoothly throughout the outage window without filter divergence, while the Y position remained bounded within the crop-row corridor.

To quantify localization drift during GNSS denial, the EKF estimated position at outage end ($t = 35$s) was compared against the simulator ground-truth trajectory using Euclidean position error:

\begin{equation}
e = \sqrt{(x_{est} - x_{gt})^2 + (y_{est} - y_{gt})^2}
\end{equation}

where $(x_{est}, y_{est})$ denotes the EKF estimated position and $(x_{gt}, y_{gt})$ denotes the simulator ground-truth position. The final positional drift after the 20-second dead-reckoning interval was 2.46~m, demonstrating that the EKF maintained bounded localization error throughout the outage window despite the absence of GNSS measurements.

\begin{table}[h]
\centering
\caption{GNSS Outage Localization Performance}
\label{tab:gnss_metrics}
\begin{tabular}{|l|c|}
\hline
\textbf{Metric} & \textbf{Value} \\ \hline
GNSS outage duration & 20 s \\ \hline
Robot forward speed & 0.3 m/s \\ \hline
Final positional drift & 2.46 m \\ \hline
Localization method & EKF dead reckoning \\ \hline
Sensor fusion inputs & GNSS + IMU + odometry \\ \hline
\end{tabular}
\end{table}

\begin{figure}[t]
    \centering
    \includegraphics[width=\columnwidth]{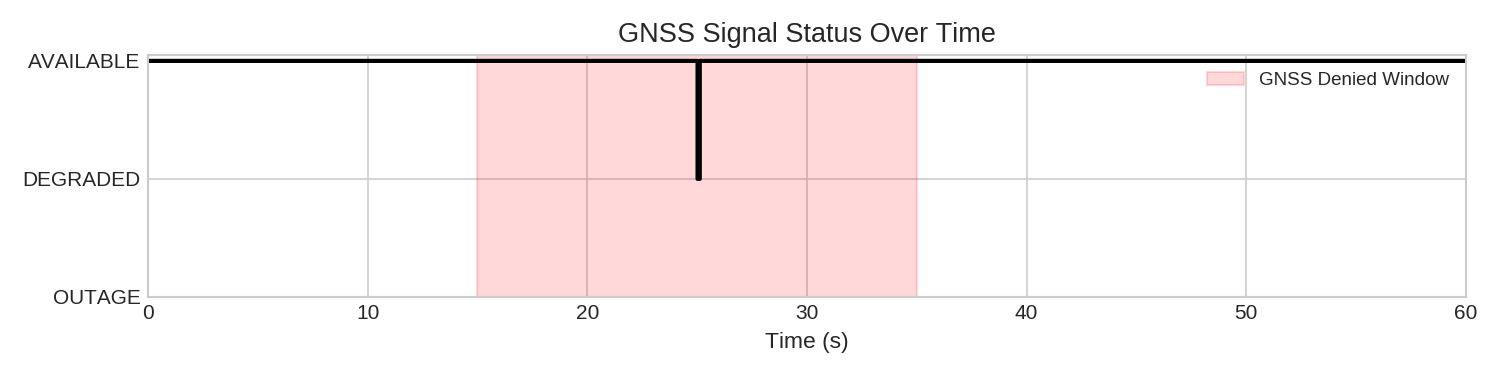}
    \caption{GNSS signal status over time. The system detects signal loss at $t \approx 17$s and re-acquires at $t \approx 35$s.}
    \label{fig:gnss_timeline}
\end{figure}

Fig.~\ref{fig:gnss_timeline} shows the GNSS signal status timeline. The system correctly detects signal loss and re-acquires without manual intervention.

\begin{figure}[t]
    \centering
    \includegraphics[width=\columnwidth]{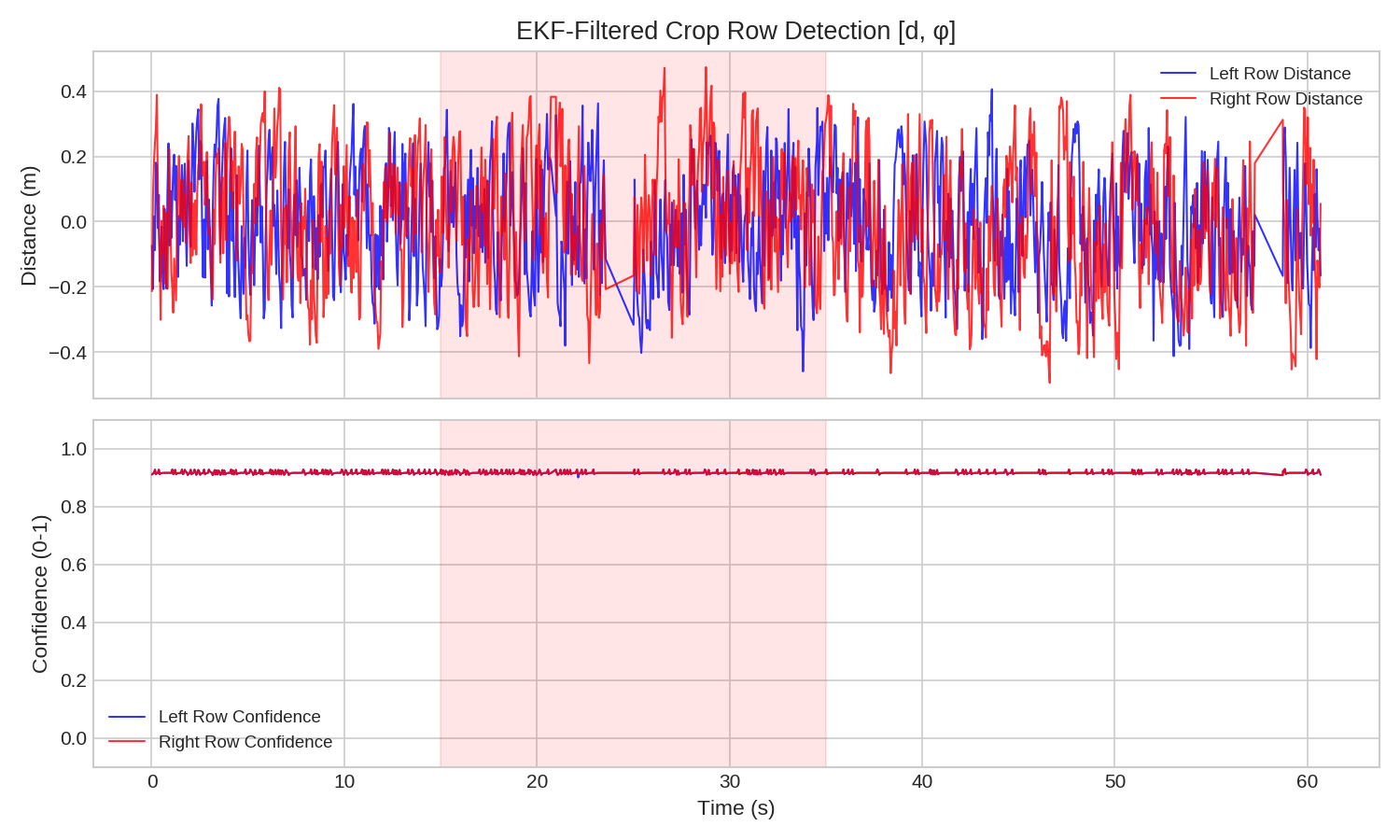}
    \caption{EKF-filtered crop row detection. Top: left and right row distances. Bottom: confidence stays above 0.9 throughout, including during GNSS outage.}
    \label{fig:croprow}
\end{figure}

Fig.~\ref{fig:croprow} shows the EKF-filtered crop row detection. The confidence metric remains above 0.9 throughout the run, including during the GNSS outage, confirming that LiDAR-based row detection operates independently of GNSS.

\begin{figure}[t]
    \centering
    \includegraphics[width=\columnwidth]{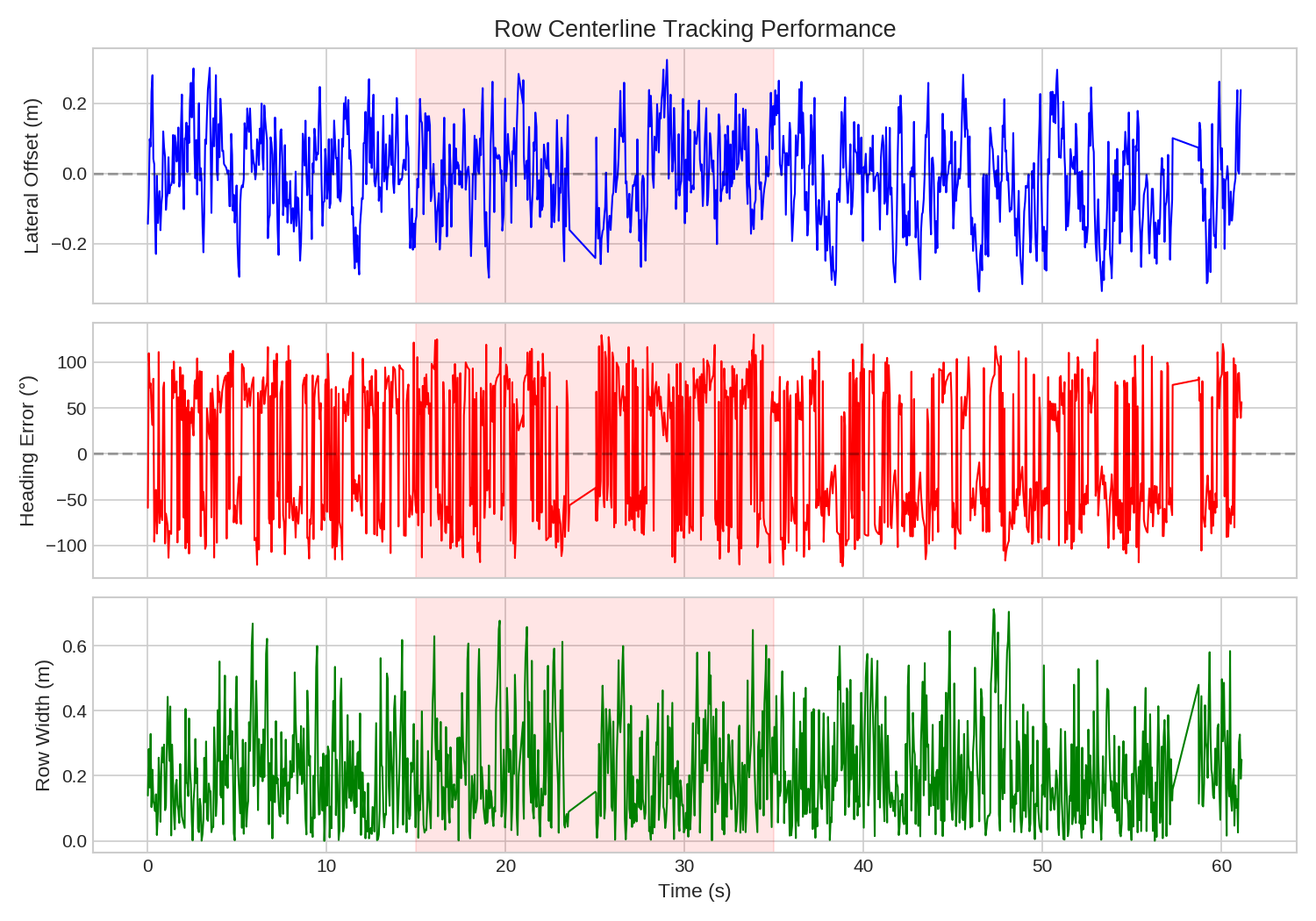}
    \caption{Row centerline tracking performance. Lateral offset (top), heading error (middle), and row width (bottom).}
    \label{fig:centerline}
\end{figure}

Fig.~\ref{fig:centerline} displays the centerline tracking. The lateral offset remains within $\pm 0.3$ m, and the pure pursuit controller maintains a nominal 0.33 m/s forward velocity with steering corrections.

\begin{figure}[t]
    \centering
    \includegraphics[width=\columnwidth]{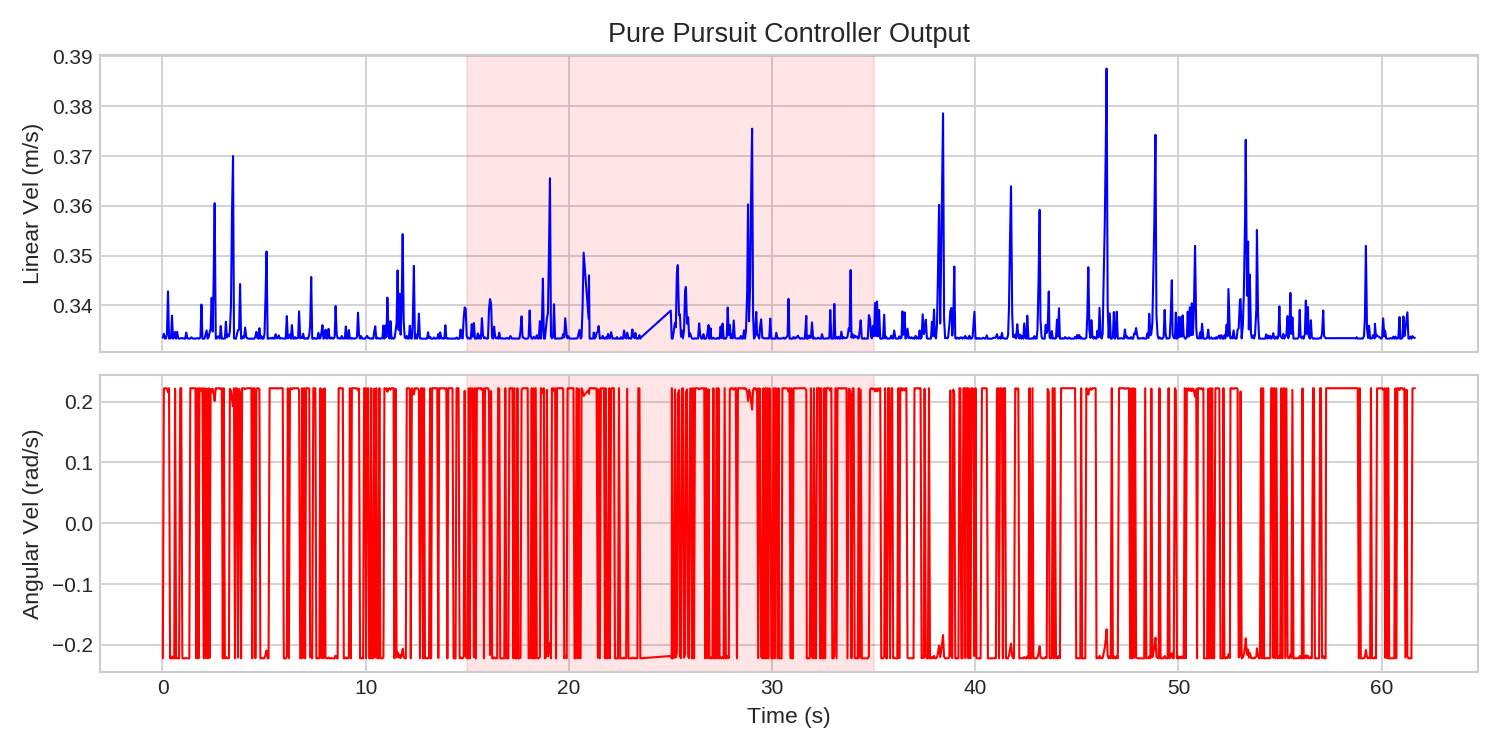}
    \caption{Pure pursuit controller output: linear velocity (top) and angular velocity (bottom) for lane following.}
    \label{fig:controller}
\end{figure}

\subsection{Integration with Team Modules}

The \texttt{/crop\_row\_state} topic provides row boundaries to the object detection module's ROI subscriber, reducing the WeedDet inference region by an estimated 30 to 50 percent. The \texttt{/row\_centerline} topic provides lateral offset and heading error for lane following. The \texttt{/localization/pose} topic provides the global pose to downstream consumers.

\section{System Integration}
\label{sec:integration}

Table~\ref{tab:summary} consolidates the key configuration parameters and observed results across the detection, navigation, and integration subsystems.

\begin{table*}[t]
\centering
\caption{AgriNav Consolidated System Results and Configuration}
\label{tab:summary}
\begin{tabular}{p{5.2cm} p{6cm} p{4.5cm}}
\toprule
\textbf{Metric} & \textbf{Value} & \textbf{Reference} \\
\midrule
\multicolumn{3}{l}{\textbf{Detection Models}} \\
\midrule
WeedDet architecture & Det-ResNet-50 + eFPN + ERetinaHead (PyTorch reimplementation) & Section~\ref{sec:weeddet} \\
WeedDet best training loss & 1.2346 over 94 epochs & Section~\ref{sec:weeddet} \\
WeedDet rice confidence range & 0.32 to 0.95 across paddy / aerial / post-flood imagery & Section~\ref{sec:weeddet} \\
Lightweight model parameters & 1.68M (5-stage backbone + 3-level FPN) & Section~\ref{sec:lightweight} \\
Lightweight training resolution & $192 \times 192$ (CPU); $512 \times 512$ GPU run pending & Section~\ref{sec:lightweight} \\
Lightweight Rice AP @ IoU 0.5 & 24.7\% & Section~\ref{sec:lightweight} \\
Lightweight Weed AP @ IoU 0.5 & 0.0\% (classification confidence to 0.995; IoU artifact) & Section~\ref{sec:lightweight} \\
Lightweight mAP @ IoU 0.5 & 12.4\% & Section~\ref{sec:lightweight} \\
Class imbalance handling & $5\times$ minority oversample + $3\times$ asymmetric loss weight & Section~\ref{sec:lightweight} \\
Per-class confidence thresholds & rice 0.25, weed 0.15 (asymmetric) & Section~\ref{sec:lightweight} \\
\midrule
\multicolumn{3}{l}{\textbf{Navigation and Localization}} \\
\midrule
Row detection method & K-means + RANSAC (100 iter, 0.05 m inlier) + per-row 2-state EKF & Section~\ref{sec:rowfollow} \\
Row detection confidence & $> 0.9$ throughout simulation, including GNSS outage & Section~\ref{sec:rowfollow} \\
Centerline lateral offset & $\pm 0.3$ m (pure pursuit controller) & Section~\ref{sec:rowfollow} \\
Nominal forward velocity & 0.33 m/s in lane-following mode & Section~\ref{sec:rowfollow} \\
Global EKF state model & 6-state CVTR fusing GNSS + IMU + wheel odometry & Section~\ref{sec:ekf} \\
Sensor rates & IMU 50 Hz, odometry 20 Hz, GNSS 1 Hz (asynchronous) & Section~\ref{sec:rowfollow} \\
GNSS outage bridged & 20 s with continuous position tracking & Section~\ref{sec:rowfollow} \\
GNSS health states & AVAILABLE / DEGRADED / OUTAGE (2 s switch threshold) & Section~\ref{sec:rowfollow} \\
Dead-reckoning $\mathbf{Q}$ inflation & Linear, $\mathbf{Q}_{DR} = \mathbf{Q}_{base}(1 + 0.5\, t_{DR})$ & Section~\ref{sec:rowfollow} \\
\midrule
\multicolumn{3}{l}{\textbf{System Integration}} \\
\midrule
LiDAR ROI compute reduction & 30 to 50\% estimated (depends on row geometry) & Section~\ref{sec:integration} \\
LiDAR-camera fusion mechanisms & 4: ROI mask, world projection, ground filter, confidence fusion & Section~\ref{sec:integration} \\
Hardware cost of fusion bridge & Zero (reuses navigation LiDAR) & Section~\ref{sec:integration} \\
\bottomrule
\end{tabular}
\end{table*}

The complete system integrates all modules over ROS. Table~\ref{tab:topics} summarizes the key inter-module topic interfaces.

\begin{table}[t]
\centering
\caption{Key ROS Topic Interfaces Between Modules}
\label{tab:topics}
\begin{tabular}{p{2.5cm}p{2cm}p{2.5cm}}
\toprule
\textbf{Topic} & \textbf{Producer} & \textbf{Consumer} \\
\midrule
/crop\_row\_state & Krish/Bilal & Benny (ROI) \\
/cmd\_vel & Krish & Vehicle Control \\
/gnss/status & Krish & All \\
\bottomrule
\end{tabular}
\end{table}

\subsection{LiDAR-Camera Bridge}

The LiDAR-camera bridge is the team's primary system-level novel contribution. It exploits the fact that LiDAR is already present on the platform for navigation. The same LiDAR data is consumed by the detection pipeline at zero additional hardware cost. Four concrete fusion mechanisms are implemented or proposed.

\textbf{Region-of-interest constraint.} The lane detection module publishes the projected LiDAR row boundaries in the camera's image plane. The detection pipeline masks the input frame to retain only the inter-row strip. This reduces inference compute by an estimated 30 to 50 percent (depending on row geometry and camera mounting angle) and removes false positives outside the rows. Weeds physically cannot grow on the row surfaces themselves, so detections in those regions are by construction false.

\textbf{World-coordinate projection.} Bounding boxes from the detection model are expressed in pixel coordinates, which is insufficient for actuation. The spray nozzle requires a coordinate in the vehicle body frame. By matching the bounding box center to the LiDAR point cloud at the corresponding pixel column, the system computes a precise ground-plane $(x, y)$ coordinate that the spray actuator consumes.

\textbf{Ground-plane removal.} LiDAR can segment points that rise above the soil ground plane. Any above-ground cluster within the inter-row zone is a candidate weed. This LiDAR-derived prior mask helps the detector ignore mud patterns and water reflections that fool RGB-only pipelines.

\textbf{Bidirectional confidence fusion.} If the detector predicts a weed at a pixel location where the LiDAR point cloud shows no physical return, the detection is downweighted as a likely false positive before reaching the spray gate. Conversely, if both modalities agree on the presence of an above-ground object, confidence is boosted. This fusion operates as an additional term in the confidence gate's geometric component.

\section{Limitations and Known Faults}
\label{sec:limitations}

This section provides an honest accounting of the components in AgriNav that are unproven, partially implemented, or known to fail under specific conditions. The team's intent is to make the gap between the current submission and a deployable system explicit, both as a roadmap for future work and as a guard against overclaiming.

\subsection{No Hardware Validation}

All results reported in this paper are from simulation or from offline inference on static imagery. No part of AgriNav has been tested on a physical tractor in a real paddy field. The Gazebo simulation provides idealized sensor noise models (Gaussian IMU noise, fixed-CEP GNSS) that do not capture multipath under wet canopy, mud-induced wheel slip, IMU vibration from tractor engine harmonics, or the camera exposure problems caused by direct overhead sun on flooded paddies. Hardware-in-the-loop testing is the single largest gap between this work and field deployment.

\subsection{Detection Module Faults}

\textbf{Issues with mAP validation for the WeedDet implementation.} A quantitative mAP evaluation was conducted using the \texttt{rice\_detection\_for\_export} validation set (1,347 images, 39,556 ground-truth boxes) against the WeedDet model trained on \texttt{Rice\_Classification.v1i} (4,041 images). Evaluation at IoU threshold 0.5 yielded mAP of less than 1 percent, and precision at IoU 0.5 was 0.4 percent across 6,000 evaluated predictions. This result is attributed primarily to cross-dataset annotation style differences: both datasets annotate rice plants, but use different box granularity, aspect ratios, and plant-grouping conventions. At a relaxed IoU threshold of 0.25, 5.6 percent of predictions overlap with ground-truth boxes, confirming that the model correctly localizes rice plant regions but with spatial offsets exceeding the standard 50 percent IoU threshold. Qualitative inference on held-out images produces consistent detections at confidence scores of 0.55 to 0.84 across paddy, aerial, and post-flood imagery, confirming that the model has learned meaningful visual features. Competitive mAP scores require retraining on a unified dataset with consistent annotation guidelines and evaluation on a held-out split from the same source distribution, which is documented as future work in Section~\ref{sec:future}.

\textbf{Lightweight model weed AP is 0.0 percent.} The lightweight CNN-FPN model achieves 0.0 percent AP on the weed class at the IoU 0.5 threshold despite producing classification confidences up to 0.995. The cause is identified (CPU-resolution training at $192 \times 192$ produces predicted boxes too small to clear the 50 percent IoU threshold) and the fix is known (GPU training at $512 \times 512$), but the fix has not been executed and the system therefore cannot pass standard detection benchmarks in its current state.

\textbf{Annotation provenance is unverified.} The training annotations were sourced from Roboflow's curated rice detection datasets. Roboflow's distribution model includes human labeling, but the labels were not authored or audited by this team. Their inter-annotator agreement statistics, labeling guidelines, and error rates are unknown. A targeted audit of a stratified sample is required before any quantitative claim about model accuracy.

\textbf{Geographic generalization is unstudied.} Both source datasets were collected in Asian rice paddies. The detector inherits the visual statistics of those environments. Transfer to other rice-growing regions, particularly West Africa and parts of South America, will require either domain-adaptive fine-tuning or region-specific data collection, neither of which has been performed.

\subsection{Localization Faults}

\textbf{Synthetic GNSS outage only.} The 20-second GNSS denial validation uses a programmed signal cut in simulation. Real GNSS degradation is gradual, intermittent, and correlated with multipath geometry; the simulated outage does not capture these characteristics.

\textbf{Linear process-noise inflation is heuristic.} The dead-reckoning process noise model $\mathbf{Q}_{DR} = \mathbf{Q}_{base} \cdot (1 + 0.5 \cdot t_{DR})$ inflates noise linearly with elapsed outage time. The 0.5 coefficient was hand-tuned to the simulated IMU drift rate. Real IMU drift is non-linear and depends on temperature, vibration, and bias instability; the linear model will under-estimate uncertainty after long outages and over-estimate it after short ones.

\textbf{No quantitative drift bound.} The paper reports that position tracking continued through the 20-second outage but does not report the absolute position error at outage end. Dead reckoning accumulates error, and a real deployment requires a documented drift envelope (centimeters per second) before the navigation stack can decide when to demand a re-localization.

\subsection{LiDAR-Camera Bridge Calibration}

The LiDAR-camera fusion bridge is presented as the team's primary system-level contribution, but its four mechanisms (ROI masking, world-coordinate projection, ground-plane filtering, bidirectional confidence fusion) all require accurate extrinsic calibration between the camera and the LiDAR. The calibration procedure is not specified in this paper, and miscalibration of even a few centimeters will cause the projected ROI mask to mis-align with the actual inter-row zone, defeating the latency and false-positive benefits.

\subsection{Safety Validation is Absent}

No controlled safety evaluation has been performed. The hardcoded rice-veto in the discrimination module has not been exercised against synthetic adversarial inputs or borderline real-world cases (young rice plants visually similar to wild grasses, water reflections that create ghost rice detections, partial-occlusion scenes where the rice canopy is interrupted). The acceptable false-spray-on-rice rate has not been formally specified, which means the system has no measurable safety target.

\section{Future Work and Real-World Deployment Path}
\label{sec:future}

This section describes how AgriNav would be extended from its current simulation-validated state to a deployed precision agriculture platform, organized by the deployment scenarios the system enables.

\subsection{Near-Term Deployment Scenarios}

\textbf{Smallholder paddy farms.} The most direct deployment target is small to medium rice farms (1 to 10 hectares) in regions where labor costs are rising and herbicide expense is a meaningful fraction of operating cost. A single AgriNav unit on a compact electric tractor platform could service such a farm in a few passes per growing season. The economic value comes from chemical reduction (estimated 60 to 80 percent less herbicide than broadcast spraying for moderate weed pressure), not labor replacement, since the tractor still requires supervision in the current state.

\textbf{Agricultural research stations.} Research stations and university experimental plots provide a controlled deployment environment with known field geometry, ground-truth weed counts, and supervised operators. AgriNav could be deployed at FGCU's own research plots or at partner institutions for site-specific weed management studies, providing the quantitative field validation that the current simulation results lack.

\textbf{Precision spray retrofits.} The detection and discrimination modules can be deployed independently of the navigation stack on existing manually-driven sprayers. The operator drives the rig as normal; the perception subsystem controls only the spray valves. This is the lowest-risk deployment path because it removes autonomy from the loop while still capturing most of the herbicide-reduction benefit.

\subsection{Required Engineering Work}

The following work items are sequenced in approximate order of dependency:

\begin{itemize}
    \item \textbf{Quantitative mAP evaluation.} Implement pycocotools-based AP@IoU0.5 and AP@IoU[0.5:0.95] evaluation against the validation and test splits for both detection models. Without this, no quantitative comparison against the published WeedDet benchmark is possible.
    \item \textbf{GPU training at production resolution.} Retrain both detection models at $512 \times 512$ resolution with full data augmentation. Expected outcome: weed AP above 30 percent, mAP above 60 percent, sufficient for deployment with the discrimination second-stage filter.
    \item \textbf{Camera-LiDAR extrinsic calibration.} Specify and validate a calibration procedure (checkerboard or AprilTag-based) that the production system can run at the start of each shift. Document the acceptable calibration error in centimeters and degrees, and add a runtime check that flags miscalibration before the spray system is enabled.
    \item \textbf{Hardware-in-the-loop testing on the Amiga robot.} Deploy on the Kantor Lab's Amiga simulation with the full WeedDet model and LiDAR ROI integration end-to-end. This exposes integration bugs that pure simulation hides.
    \item \textbf{Real GNSS denial characterization.} Replace the synthetic 20-second outage with logged GNSS data from real paddy environments under different canopy conditions (early growth, mid-season, late-season closure). Re-tune the dead-reckoning process noise model on real drift profiles, then characterize the position drift envelope as a function of outage duration.
    \item \textbf{Embedded deployment.} Quantize both detection models to INT8 via TensorRT and benchmark on the NVIDIA Jetson AGX Orin. Characterize the accuracy loss from quantization and the latency budget at the target frame rate.
    \item \textbf{Controlled safety phase.} Build a synthetic adversarial test suite for the rice-veto (young rice plants, water reflections, partial occlusion, lighting extremes) and a documented borderline-case dataset. Specify and measure the acceptable false-spray-on-rice rate before any field trial.
\end{itemize}

\subsection{Longer-Term Research Directions}

\textbf{Multi-spectral and thermal sensing.} The current system uses a single RGB camera. Adding a near-infrared band would substantially improve plant-versus-soil discrimination under variable lighting, and a thermal channel would help detect water stress or weed sub-surface roots. Both modalities integrate naturally with the existing detection pipeline as additional input channels.

\textbf{Cross-region domain adaptation.} Adapting to new geographic regions could be done with a small labeled set from a target region (West African rice, Latin American rice) plus a larger unlabeled set, fine-tuning the detection model without requiring full re-annotation.

\textbf{Fleet-level coordination.} A single-tractor system is the right unit for smallholder deployment. For larger operations, multiple AgriNav units could share row-detection and weed-density maps over a low-bandwidth field network, allowing the fleet to load-balance and avoid double-spraying. This requires no changes to the per-tractor stack but would benefit from a centralized field-state representation.

\textbf{Long-term field learning.} A deployed AgriNav unit could log its detections, confidence scores, and false-positive corrections to a per-farm dataset that fine-tunes the detection model over multiple growing seasons. This compounds: each season's deployment improves the next season's accuracy without requiring central data collection.

\section{Conclusion}
\label{sec:conclusion}

This paper presented the design of an autonomous agricultural tractor system integrating LiDAR navigation and vision-based weed detection. The lane detection and localization subsystem demonstrated robust crop row tracking with confidence above 0.9 and successful GNSS outage bridging through a 20-second signal denial period. The custom WeedDet implementation trained for 94 epochs reached a best training loss of approximately 1.2346. Cross-dataset quantitative evaluation confirmed partial localization capability (5.6 percent precision at IoU 0.25) with qualitative confidence scores of 0.55 to 0.84 across diverse field conditions. Full mAP evaluation on a same-distribution held-out split remains as immediate future work. The inverted-logic discrimination module addresses the asymmetric-cost problem through a hardcoded confidence-gate veto on the rice class.

The team's primary system-level contribution is a four-mechanism LiDAR-camera fusion bridge that exploits the navigation LiDAR for ROI constraints, world-coordinate projection, ground-plane filtering, and bidirectional confidence fusion. Estimated detection inference compute is reduced by 30 to 50 percent through the ROI constraint alone.

Section~\ref{sec:limitations} documents the gap between the current simulation-validated state and a deployable system, and Section~\ref{sec:future} sequences the engineering work required to close that gap across three concrete deployment scenarios: smallholder paddy farms, agricultural research stations, and precision spray retrofits on existing manually-driven equipment.

\end{document}